\documentclass{article}

\usepackage[preprint,nonatbib]{neurips_2019}

\usepackage{multirow}
\usepackage[utf8]{inputenc} 
\usepackage[T1]{fontenc}    
\usepackage{hyperref}       
\usepackage{url}            
\usepackage{booktabs}       
\usepackage{amsfonts}       
\usepackage{nicefrac}       
\usepackage{microtype}      
\usepackage{graphicx}
\usepackage{wrapfig}
\newlength\savewidth
\newcommand\whline{\noalign{\global\savewidth\arrayrulewidth
                            \global\arrayrulewidth 1pt}%
                   \hline
                   \noalign{\global\arrayrulewidth\savewidth}}
                   
\newcommand\shline{\noalign{\global\savewidth\arrayrulewidth
                            \global\arrayrulewidth 0.5pt}%
                   \hline
                   \noalign{\global\arrayrulewidth\savewidth}}

\title{Weakly-supervised Compositional Feature Aggregation for Few-shot Recognition}

\author
{
Ping Hu \\
Boston University\\
\texttt{pinghu@bu.edu} \\
\And
Ximeng Sun \\
Boston University\\
\texttt{pinghu@bu.edu} \\
\And
Kate Saenko \\
Boston University\\
\texttt{pinghu@bu.edu} \\
\And
Stan Sclaroff \\
Boston University\\
\texttt{pinghu@bu.edu} \\
}

\begin{document}

\maketitle

\begin{abstract}

Learning from a few examples is a challenging task for machine learning. While recent progress has been made for this problem, most of the existing methods ignore the compositionality in visual concept representation (e.g. objects are built from parts or composed of semantic attributes), which is key to the human ability to easily learn from a small number of examples. To enhance the few-shot learning models with compositionality, in this paper we present the simple yet powerful Compositional Feature Aggregation (CFA) module as a weakly-supervised regularization for deep networks. Given the deep feature maps extracted from the input, our CFA module first disentangles the feature space into disjoint semantic subspaces that model different attributes, and then bilinearly aggregates the local features within each of these subspaces. CFA explicitly regularizes the representation with both semantic and spatial compositionality to produce discriminative representations for few-shot recognition tasks. Moreover, our method does not need any supervision for attributes and object parts during training, thus can be conveniently plugged into existing models for end-to-end optimization while keeping the model size and computation cost nearly the same. Extensive experiments on few-shot image classification and action recognition tasks demonstrate that our method provides substantial improvements over recent state-of-the-art methods.

\end{abstract}

\section{Introduction}


The human visual system has a remarkable ability to efficiently learn semantic concepts from just one or a few examples~\cite{lake2015human}. 
To achieve a similar ability with machine learning, Few-Shot Learning (FSL) has emerged as an important research topic in recent years~\cite{WangFSLSurvey,fei2006one,fink2005object,Hariharan_2017_ICCV,Xu_2017_CVPR,Wang_2017_CVPR,xing2019adaptive,yang2018one,rakelly2018conditional,xu2018similarity,kang2018few}.  Given a small set of labeled examples (\textit{support set}) of  novel object categories (\textit{novel classes}), FSL aims to apply a model trained on the known object classes (\textit{base classes}) to classify the unlabeled samples (\textit{query set}) from the novel classes. To tackle this problem, meta-learning based models~\cite{finn2017model,ravi2016optimization,Qiao_2018_CVPR,Gidaris_2018_CVPR} propose to train a meta-learner that can be quickly adapted to the new recognition tasks for the novel object categories;  feature-hallucination approaches~\cite{Wang_2018_CVPR,Hariharan_2017_ICCV} learn to generalize the base classes' distribution to augment samples in the query set. Compared to these models with sophisticated frameworks and protocols, a more straightforward yet effective approach employs metric-learning based models~\cite{vinyals2016matching,snell2017prototypical,Sung_2018_CVPR,Qi_2018_CVPR} that classify samples in the query set based on the metric distance to the labelled examples in the support set.

\begin{figure}
\centering
\includegraphics[height=4.6cm] {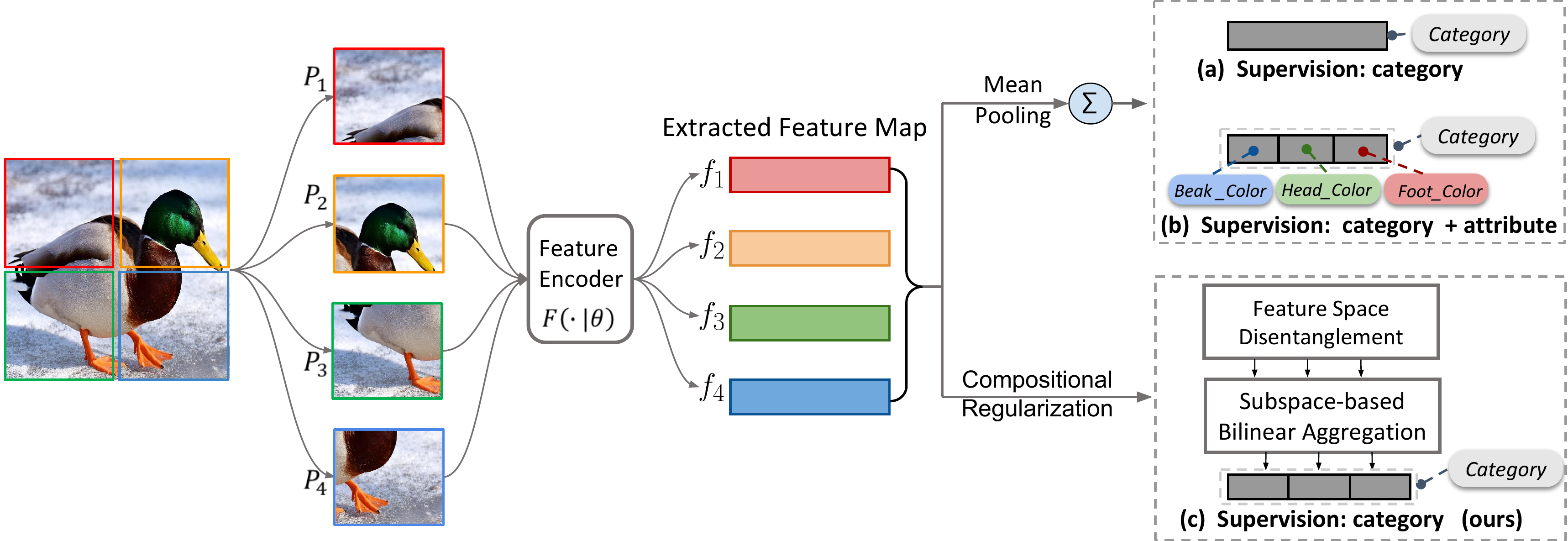}
\vspace{-0.2cm}
\caption{\small Deep CNNs map local images parts (e.g. $P_1 \sim P_4$)  into  feature vectors (e.g. $f_1 \sim f_4$). (a) Spatially mean/max pooling over the feature maps loses spatial compositional information. (b) Further training to learn semantic compositionality requires annotations for attributes. (c) Our CFA module first disentangles the feature space, then bilinearly aggregates feature maps in each subspace, thus successfully enforcing both spatial and semantic compositionality in the network in a weakly supervised way.}
\label{fig0}
\vspace{-0.5cm}
\end{figure}

 Although the aforementioned methods have achieved significant results when combined with deep CNNs' superior learning and generalization capabilities, they may still suffer from the inherent limitations of FSL tasks~\cite{WangFSLSurvey}. The size of the \textit{support set} is typically too small to reliably generalize the model learned on base categories to novel classes. To alleviate this issue, one possible approach is to take into consideration the compositionality of concept representation (e.g. the fact that objects are built from parts and composed of semantic attributes).  Compositionality plays a key role in the human visual system, as it represents novel concepts as known primitives, which helps us learn efficiently from a few examples~\cite{hoffman1984parts,biederman1987recognition,marr1978representation}. Inspired by these findings, models in~\cite{tokmakov2018learning,andreas2019measuring} learn with attribute-level annotations to encourage the deep features to encode semantic compositionality. However,  as shown in Fig.~\ref{fig0}, such methods need to predefine a fixed set of attributes and rely on attribute-level annotations for training, which may be sub-optimal and limit the range of applications. Moreover, methods including~\cite{tokmakov2018learning,andreas2019measuring,vinyals2016matching,snell2017prototypical,Sung_2018_CVPR,Qi_2018_CVPR,Wang_2018_CVPR} apply mean/max pooling over feature maps to produce image-level representations, thus losing the objects' spatial compositionality which is important for visual understanding.

In order to effectively impose both the spatial and semantic compositionality to enhance few-shot learning performance, in this work we propose the Compositional Feature Aggregation (CFA) module as a weakly-supervised module for end-to-end optimization.  Given the feature maps extracted from the input, at first we explicitly  disentangle the feature space into independent semantic subspaces to encourage semantic compositionality. Then, to further impose spatial compositionality in each of the semantic subspaces, rather than simply applying mean/max pooling, we aggregate the sub-feature maps via bilinear aggregation to extract second-order statistics and capture translation-invariant spatial structure.  Finally, we concatenate the aggregated feature vectors from all the subspaces and use it as the final descriptor. The proposed CFA module explicitly imposes semantic and spatial compositionality to help models focus on generalizing semantic knowledge at the attribute-level and object-part-level rather than at a holistic level, thus
improving learning and generalization.
Moreover,  CFA imposes the compositionality onto deep features in an weakly-supervised way and does not need any annotations for attributes and object parts for training. 
 
To summarize, this paper makes the following contributions.   (i) We propose to explicitly impose both semantic and spatial compositionality in the form of weakly supervised regularization for deep networks to improve generalization in few-shot recognition tasks. (ii) We propose CFA as a convenient plugable module for end-to-end optimization
without requiring annotations for semantic attributes or object parts. (iii) We evaluate our method with extensive experiments for few-shot image classification and action recognition tasks. The state-of-the-art performance validates the our method's effectiveness for few-shot recognition tasks.

\section{Related Work}
\noindent\textbf{Few-shot Learning.} Few-shot learning methods aim to classify new categories based on limited supervision information. Recent trends for this task can be roughly grouped into three types: meta-learning based approaches~\cite{munkhdalai2017meta,mishra2017simple}, feature-hallucination based methods~\cite{Hariharan_2017_ICCV,Wang_2018_CVPR}, and metric-learning based models~\cite{kochsiamese}. The meta-learning based approaches aim to learn a "meta-learner" that provides proper initialization~\cite{finn2017model} or weight updates~\cite{ravi2016optimization} for models to quickly adapt to novel tasks with few training examples. The feature-hallucination based methods adopt generators to learn to transfer data distributions~\cite{Hariharan_2017_ICCV,Wang_2018_CVPR} or visual styles~\cite{antoniou2018data} to augment the novel examples. The metric-learning based models learn to encode and compare features  such that samples of the same category show higher similarity than those of different categories, where the similarity can be evaluated with cosine similarity~\cite{vinyals2016matching}, euclidean distance~\cite{snell2017prototypical}, deep relation module~\cite{Sung_2018_CVPR}, and graph neural networks~\cite{garcia2017few,Guo_2018_ECCV}. Qi et al.~\cite{Qi_2018_CVPR} utilize the features of novel classes as class prototypes to extend the weight matrix of the final classification layer, so that the networks can dynamically process both base and novel classes. Similarly, Gidaris et al.~\cite{Gidaris_2018_CVPR} and Qiao et al.~\cite{Qiao_2018_CVPR} learn to predict the weights of final classification layer for novel classes. Inspired by the compositionality in human's visual perception, Tokmakov et al.~\cite{tokmakov2018learning} proposed to explicitly learn semantic compositional representations for few-shot image learning. However,~\cite{tokmakov2018learning} requires attribute-level annotations during training, and applies mean pooling operation that loses discriminative information contained in the object parts' spatial structure. In contrast, our method successfully imposes both spatial and semantic compositionality in an weakly supervised way without the need for supervision of semantic attribute or object parts, and can conveniently learn end-to-end to produce discriminative representations.          

\noindent\textbf{Compositional Representation.} Compositionality plays a key role in the human vision system, as it allows to represent novel concepts as knowing primitives so that to learn efficiently from a few examples~\cite{hoffman1984parts,biederman1987recognition,marr1978representation}. To exploit this feature to enhance deep neural networks, Andreas et al.~\cite{andreas2019measuring}  and Tokmakov et al.~\cite{tokmakov2018learning} utilize attribute annotations to learn deep embedding for compositional feature, which is a sum of encodings of the attributes of the inputs. Misra et al.~\cite{misra2017red} train classifiers for different attributes and combine them to represent novel concepts. One limitation for these methods is that they apply mean/max pooling operations over the feature maps thus losing the spatial compsitionality of visual concepts and may result in less discriminative representations. Stone et al.~\cite{stone2017teaching} address the  spatial compsitionality by constraining the object parts to be independent in the representation space.  However, all these methods rely on annotations for attributes or object parts.

\noindent\textbf{Bilinear Feature Aggregation.} Bilinear models were propose in~\cite{tenenbaum2000separating} to model two-factor variations like  “style” and “content” for images. To improve the image recognition performance with richer spatial structure, bilinear models has been utilized to model the variations arising out of appearance and part locations~\cite{lin2015bilinear,girdhar2017attentional,chen20182}. Comparing to mean/max pooling operations~\cite{goodfellow2016deep} that extract first-order statistics, the bilinear aggregation models~\cite{lin2015bilinear,girdhar2017attentional} compute second-order statistics to preserve more complex relations.  Lin et al.~\cite{lin2015bilinear} shows that bilinear model also generalizes to orderless second-order pooling techniques like VLAD~\cite{arandjelovic2016netvlad,jegou2011aggregating} and Fisher Vector~\cite{cimpoi2015deep}. In our method, in order to retain richer spatial compositional  information when aggregating feature maps, we build our method on the NetVLAD~\cite{arandjelovic2016netvlad,girdhar2017actionvlad}, which is a differentiable version of VLAD~\cite{jegou2011aggregating}, and extend it with semantic compositionality to enhance performance for few-shot learning.

\section{Compositional Feature Aggregation}
\begin{figure}
\centering
\includegraphics[height=4.2cm] {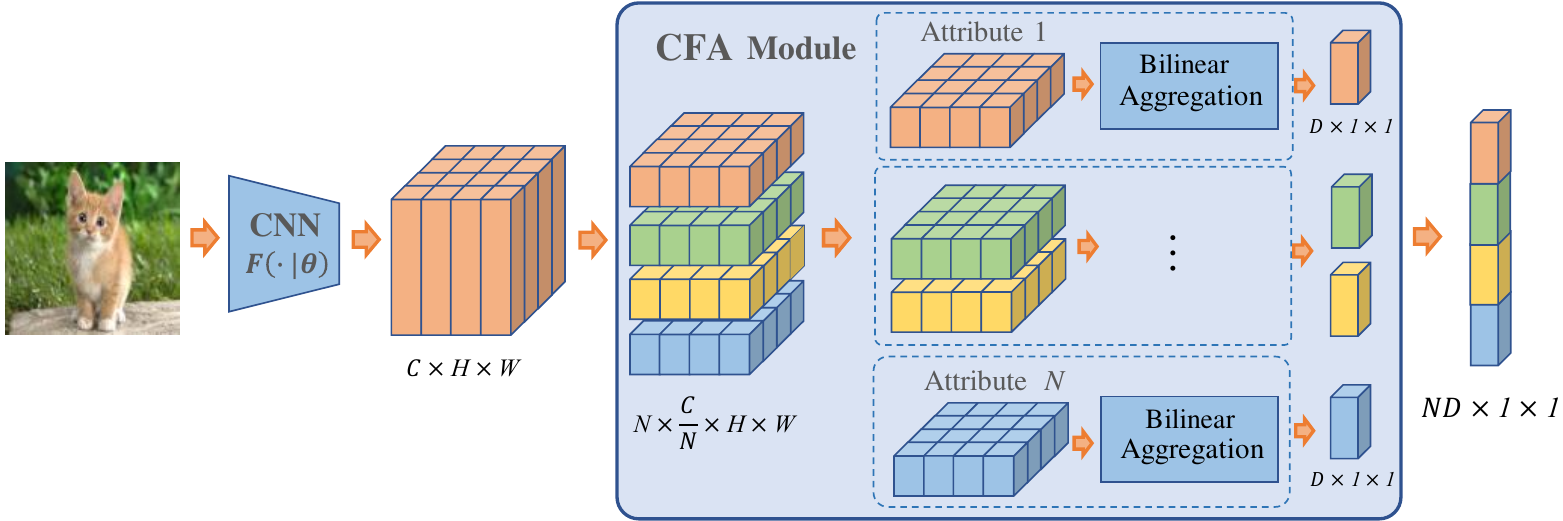}
\vspace{-0.25cm}
\caption{\small The proposed Compositional Feature Aggregation (CFA). The $C\times H\times W$ feature maps are extracted from an input image, and then split into $N$ parts along the channel dimension to be a group of  $\frac{C}{N}$-dimension semantic subspaces. Within each of the subspace, the  $\frac{C}{N}\times H\times W$ sub-feature map are bilinearly aggregated into a $D$-dimension vector ($D$ is decided by the aggregating method used) so that to extract richer spatial compositional information. Finally, the sub-feature vectors produced in all the semantic subspaces are concatenated to be a $N\cdot D$-dimension final representation,  and further input to a cosine similarity based nearest neighbor classifier.}
\label{fig1}
\vspace{-0.6cm}
\end{figure}

 As shown in Fig.~\ref{fig1}, our CFA module is a trainable module that can be conveniently plugged into standard deep CNNs to learn to aggregate image-level compositional semantic representations. By decomposing the semantic feature space into subspaces and bilinearly aggregating features in each of them, we impose both semantic and spatial compositionality in a  weakly-supervised way without requiring annotations of semantic attributes and object parts for training. 
 
 
 
\subsection{Problem Setting for FSL}
 We follow the  $X$-way $Y$-shot protocol adopted in recent few-shot learning methods~\cite{vinyals2016matching,snell2017prototypical,Sung_2018_CVPR,chen2019closer}. Formally, an $X$-way $Y$-shot learning task involves three sets of data: a training set $\textbf{T}_{b}$ containing labelled samples from the base classes; a support set $\textbf{S}_{n}$ consisting of $X$ novel classes with $Y$ labelled examples for each; and a query set $\textbf{Q}_{n}$ composed by unlabelled examples from the same $X$ novel classes. Usually, the amount of samples for each class in $\textbf{T}_{b}$ is much larger than $Y$. To learn an effective embedding for the few-shot learning task, we adopt \textit{episode}-based training that utilizes training samples to mimic the target task that classifies samples in $\textbf{Q}_{n}$ conditioned on  $\textbf{S}_{n}$. At each training iteration we randomly sample $X$ classes with  $Y$ labelled samples for each from $\textbf{T}_{b}$ to play as a support set $\textbf{S}_{b}$, and a fraction of the remaining samples in the same $X$ base classes are selected as the query set $\textbf{Q}_{b}$. The objective is to train a nearest neighbor based classifier $\textbf{M}$ to minimize the $X$-way prediction loss. In the \textit{episode} testing stage after training, we apply $\textbf{M}$ to perform nearest neighbor searching over $\textbf{S}_{n}$ to classify the query samples in $\textbf{Q}_{n}$.
 
\subsection{Semantic Decomposition}

 One of the key factors for human vision's superior ability to learn from few examples is the semantic and spatial compositionality in concept representation~\cite{hoffman1984parts,biederman1987recognition,marr1978representation}. In order to conveniently learn to produce a compositional representation without the need of extra annotations, inspired by \textit{Group Convolution}~\cite{krizhevsky2012imagenet,zhang2017interleaved}, we explicitly decompose the feature space into independent subspaces in order to regularize the deep representation with semantic compositionality. Given a deep CNN $F(\cdot|\theta)$ that maps an image patch into a $C$-dimension vector, we uniformly divide the vector into a predefined number $N$ disjoint groups along the channel dimension. Each of these sub-vectors has $\frac{C}{N}$ channels and corresponds to one semantic subspace. By further explicitly imposing spatial compositionality within each of the semantic subspace (details in the next subsection), we can regularize $F(\cdot|\theta)$ to focus on generalizing semantic knowledge at the attribute-level and part-level rather than at the holistic instance-level, thus reducing the difficulties in learning. Since our method doesn't use attribute-level annotations, the manually defined $N$ subspace may correspond to semantic attributes that are not as meaningful for humans as the predefined attributes in~\cite{andreas2019measuring,tokmakov2018learning}. However, since our model can be conveniently optimized end-to-end, the learnt attributes can be better adapted to the task. It is also possible to explore other methods to group the feature channels rather than evenly dividing into $N$ parts, yet this is beyond the scope of this paper and we leave it for future research.

\subsection{Bilinear Aggregation in Semantic Subspace}
 After decomposing the feature space into the $N$ semantic subspaces, the feature maps encoded by $F(\cdot |\theta)$ are divided into $N$ groups of sub-feature maps with the same spatial structure but correspond to different semantic attributes. In order to convert these feature maps into a fixed-length representation vector, a normal practice as in~\cite{tokmakov2018learning,vinyals2016matching,snell2017prototypical,Sung_2018_CVPR} is spatially mean/max pooling over the feature map. Yet the mean/max pooling operations will lose the spatial compositionality information of object parts in the input image, leading to sub-optimal and less discriminative representations for few-shot recognition tasks. Another straightforward choice can be directly flatten the feature maps, which will keep the exact spatial structure of the input image. However, we found that this will drastically decrease the performance, because directly flatten the feature maps is not translation-invariant yet objects from the same category may show different spatial layouts.

\begin{wrapfigure}{r}{0.53\textwidth}
\includegraphics[width=0.53\textwidth] {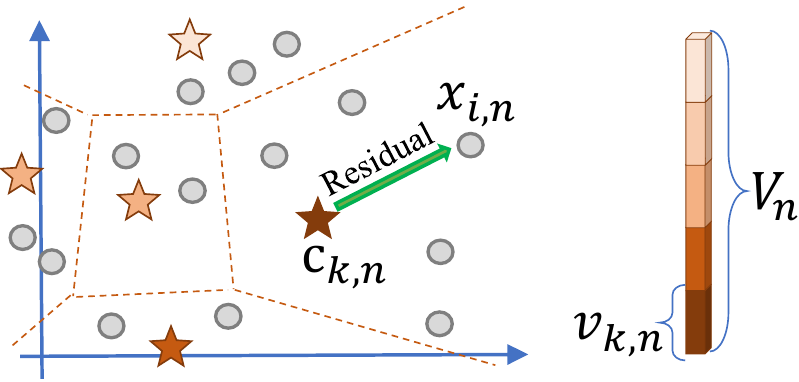}
\vspace{-0.6cm}
\caption{ An example of local aggregation in a semantic subspace $n$. The residual vectors between local features $x_{i,n}$ and the corresponding attribute prototype $c_{k,n}$ are summed within each cell to obtain $ v_{k,n}$, and then concatenated together to form $V_n$.}
\label{fig1-1}
\vspace{-0.2cm}
\end{wrapfigure}

To effective retain the spatial compositionality, we propose to bilinearly aggregate local features in each of the semantic subspace. As shown in~\cite{lin2015bilinear,jegou2011aggregating,arandjelovic2013all}, the Vector of Locally Aggregated Descriptor (VLAD)~\cite{jegou2011aggregating} as a generalized bilinear model is able to aggregate feature maps in a translation-invariant way  without losing spatial compositional information. Thus we built our aggregation model on the NetVLAD~\cite{arandjelovic2016netvlad}, which is a differentiable version of VLAD. Given a $H\times W$ feature maps, consider $x_{i,n}\in \mathcal{R}^{\frac{C}{N}}$ to be the $\frac{C}{N}$-dimension feature at spatial location $i\in \{1,...,HW\}$ in semantic subspace $n\in \{1,...,N\}$. We learn to divide the $\frac{C}{N}$-dimension semantic subspace $n$ into $K$ cells via $K$ cluster centers ("semantic prototypes")   $\{c_{k,n} | k=1,..,K \}$. Each local semantic sub-feature $x_{i,n}$ is then assigned to its nearest center and the residual vector $x_{i,n}-c_{k,n}$ is recorded. For each of the cells in semantic subspace $n$, the residual vectors are then summed spatially as, 
 \small
 \begin{equation}
     v_{k,n} = \sum_{i=1}^{HW}\frac{e^{-\alpha ||x_{i,n}-c_{k,n}||^2}}{\sum_{k'}e^{-\alpha ||x_{i,n}-c_{k',n}||^2}} (x_{i,n}-c_{k,n})
 \label{eq0}
 \end{equation} 
 \normalsize
 where the $\alpha$ is always set to be high (100 in our experiments) to achieve the effect of hard assignment, and $K$=32 as suggested in~\cite{arandjelovic2016netvlad,jegou2011aggregating}. The $v_{k,n}$ is a $\frac{C}{N}$-dimension vector that describes the distribution of the input object's local parts in the cell with the $k$-th semantic prototype of the $n$-th semantic subspace. An illustration is shown in Fig.~\ref{fig1-1}. Comparing to the mean/max pooling that pool features over all the entire features space, the local aggregation method can be seen to pool features within cells of each  semantic prototypes, thus retaining richer spatial compositional information.

 In the case of few-shot recognition task, we may have $Y>1$ labeled examples as support for each novel class. Based on Equation~\ref{eq0}, given $x_{i,n}^t$ as $x_{i,n}$ from the $t$-th sample in the support set, the information in multiple samples can be conveniently aggregated as, 
  \small
  \begin{equation}
     v_{k,n} =\frac{1}{Y} \sum_{t=1}^{Y}\sum_{i=1}^{HW}\frac{e^{-\alpha ||x_{i,n}^t-c_{k,n}||^2}}{\sum_{k'}e^{-\alpha ||x_{i,n}^t-c_{k',n}||^2}} (x_{i,n}^t-c_{k,n})
 \label{eq1}
 \end{equation}
 \normalsize
 By stacking all the $\{v_{k,n}|k=1,...,K\}$ together, we get an $\frac{CK}{N}$-dimension descriptor for the input image in $n$-th semantic subspace, which is $V_{n} = [v_{1,n}; v_{2,n}; ...; v_{K,n}]$.  Further, we concatenate the descriptors in different semantic subspaces together to be the $CK$-dimension overall representation,
  \begin{equation}
     I = [V_{1}; V_{2}; ...; V_{N}] 
 \label{eq3}
 \end{equation}
 
 The image-level representation $I$ is finally L2-normalized as suggested in~\cite{arandjelovic2016netvlad} to be the our compositional aggregated feature. With such a design, our CFA module is able to explicitly impose spatial compositionality information as regularization for deep representations. Moreover, different from mean/max pooling operations that aggregate both foreground and background features equally, the above locality based feature aggregation also helps to highlight the similar contents among images and suppress the influence of background~\cite{arandjelovic2016netvlad,jegou2011aggregating}, which helps to produce more discriminative representations for few-shot recognition. As all the parameters of our CFA module are differentiable so that the proposed module can be conveniently plugged into other deep CNNs for end-to-end optimization. It should be noted that the aggregation method we use is based on NetVLAD~\cite{arandjelovic2016netvlad}, yet we novelly extend it with semantic compositionality to tackle the task of few-shot recognition.

\subsection{CFA for Few-shot Recognition}
As our CFA module naturally aggregate multiple support examples into a single L2-normalized representation vector, we adopt the cosine-similarity-based nearest neighbor classifier,
 \small
 \begin{equation}
    \hat{l} = \sum_{i=1}^Y\left (\frac{e^{d(I_i,\hat{I})}}{\sum_{j=1}^Ye^{d(I_j, \hat{I})}} \right) l_i
 \label{eq4}
 \end{equation}
 \normalsize
where $d(\cdot)$ is the cosine similarity, $I_i, i\in\{1,..,Y\}$ is the overall representation vector for the $i$-th category,  $\hat{I}$ is the representation vector for the query sample, $l_i,  i\in\{1,..,Y\}$ is the class label for $i$-th category, and $\hat{l}$ is the predicted label for the query sample. Given the groundtruth label $l_{gt}$ for the query sample, we adopt the cross entropy loss $L_{cls} = - l_{gt}\cdot log(\hat{l})$ as the objective for classification.

Furthermore, in the proposed CFA module, the $\frac{C}{N}$-dimension semantic prototypes $\{c_{k,n} | k=1,..,K; n=1,...,N \}$ are important learnable parameters as the local features are grouped and aggregated over them. To avoid learning trivial results for these parameters during the end-to-end training, we add a regularization to the loss function to enforce the orthogonality between semantic prototypes  within each attribute subspace. As a result, the final loss function for training is,
 \small
 \begin{equation}
    L_{CFA} = - l_{gt}\cdot log(\hat{l}) + \gamma \sum_{n=1}^N |(c_{k,n})(c_{k,n})^T - Iden(K)|
 \label{eq5}
 \end{equation}
 \normalsize
where $\gamma$ is a weight for orthogonality constraint, and $Iden(K)$ is a $K\times K$ identity matrix. 

\section{Experiments}
\vspace{-5pt}
We evaluate our method on two few-shot recognition scenarios: image classification, and action recognition. All the experiments are implemented with Pytorch on a Nvidia Titan Xp GPU card.

\vspace{-5pt}
\subsection{Image Classification}
\vspace{-5pt}
\noindent\textbf{Dataset.} The miniImagenet~\cite{vinyals2016matching} is a popular dataset for evaluating few-shot learning models. It contains 100 classes with 600 images for each from the ImageNet dataset~\cite{deng2009imagenet}. We follow the data splits adopted by~\cite{chen2019closer,ravi2016optimization} with 64 base, 16 validation, and 20 novel categories. Another dataset we use is the CUB dataset~\cite{wah2011caltech} which is a fine-grained bird species dataset composed by 11,788 images for 200 classes. We evaluate on this dataset with 64, 16, and 20 classes for training, validation, and testing respectively as in~\cite{chen2019closer, hilliard2018few}. 

\noindent\textbf{Implementation.} We utilized ResNet-18~\cite{he2016deep} as feature encoder $F(\cdot|\theta)$, and randomly initialize the parameters before training on each dataset. To effectively train with our CFA module, we first pretrain the feature encoder as a normal classifier on the base classes for 30,000 iterations, and then perform jointly episode training with loss function in Equation (\ref{eq5}) for another 30,000 iterations. The validation set is used to select the iterations of best accuracy. In each testing episode of a $Y$-shot recognition task, we randomly select 5 classes from the testing set. For each class, $Y$ labelled examples and 16 unlabelled examples are selected as support set and query set respectively. To evaluate the performance, we perform 600 testing episodes and compute the averaged accuracy with 95$\%$ confidence intervals as final results.  We also implement and evaluate several recent state-of-the-art few-shot learning methods~\cite{vinyals2016matching,snell2017prototypical,Sung_2018_CVPR,finn2017model,girdhar2017attentional} in the same setting for fair comparisons. All these methods are optimized using Adam optimizer with initial learning rate 0.001 and batchsize 16 for 60,000 iterations.

  \begin{table}
  \small
  \centering
    \begin{tabular}{ccccccc}
     \whline
        \vspace{0.015in}
        &\footnotesize{CFA(\footnotesize{N=64})} &\footnotesize{CFA(\footnotesize{N=1})} &\footnotesize{MatchingNet}\footnotesize{\cite{vinyals2016matching}}    &ProtoNet\footnotesize{\cite{snell2017prototypical}} &\footnotesize{RelationNet}\footnotesize{\cite{Sung_2018_CVPR}} &MAML\footnotesize{\cite{finn2017model}}\\ 
     \shline
      1-shot &\textbf{58.5 $\pm$ 0.8} &54.9 $\pm$ 0.9 &53.0 $\pm$ 0.8 &54.1 $\pm$ 0.9 &52.6 $\pm$ 0.9 &50.2 $\pm$  1.0\\ 
      3-shot &\textbf{70.9 $\pm$ 0.7} &68.6 $\pm$ 0.8 &62.2 $\pm$ 0.7 &68.4 $\pm$ 0.7 &63.6 $\pm$ 0.7 &63.0 $\pm$  0.8\\ 
      5-shot &\textbf{76.6 $\pm$ 0.6} &74.1 $\pm$ 0.7 &68.4 $\pm$ 0.7 &73.8 $\pm$ 0.7 &68.8 $\pm$ 0.7 &64.9 $\pm$  0.7\\ 
     \whline
     \end{tabular}
    \vspace{-0.1cm}
    \caption{The 5-way few-shot image classification accuracy (in $\%$) on the miniImagenet dataset. The ``$N$'' is the number of semantic subspaces. }
   \label{table1}
    \vspace{-0.5cm}
  \end{table}
  \normalsize
 
  \begin{table}
  \small
  \centering
    \begin{tabular}{ccccccc}
     \whline
        \vspace{0.015in}
        &\footnotesize{CFA(\footnotesize{N=64})} &\footnotesize{CFA(\footnotesize{N=1})} &\footnotesize{MatchingNet}\footnotesize{\cite{vinyals2016matching}}    &ProtoNet\footnotesize{\cite{snell2017prototypical}} &\footnotesize{RelationNet}\footnotesize{\cite{Sung_2018_CVPR}} &MAML\footnotesize{\cite{finn2017model}}\\ 
     \shline
      1-shot &\textbf{73.9 $\pm$ 0.8} 	&70.4 $\pm$ 1.0 &72.2$\pm$ 0.9 	&71.2 $\pm$ 0.8 		&67.8 $\pm$ 1.1 &68.2 $\pm$  1.0\\ 
      3-shot &\textbf{84.3 $\pm$ 0.6} 	&80.5 $\pm$ 0.6 &81.6 $\pm$ 0.6 &83.2 $\pm$ 0.6 		&80.1 $\pm$ 0.7 &79.0 $\pm$  0.7\\ 
      5-shot &86.8 $\pm$ 0.5 		    &82.7 $\pm$ 0.6 &83.7 $\pm$ 0.5	&\textbf{87.1 $\pm$ 0.5}&82.8 $\pm$ 0.6 &83.0 $\pm$  0.6\\ 
     \whline
     \end{tabular}
    \vspace{-0.1cm}
    \caption{The 5-way few-shot image classification accuracy (in $\%$) on the CUB dataset. }
   \label{table2}
    \vspace{-0.75cm}
  \end{table}
  \normalsize

\noindent\textbf{Results.} We compare with recent state-of-the-art models including MatchingNet~\cite{vinyals2016matching}, ProtoNet~\cite{snell2017prototypical}, RelationNet~\cite{Sung_2018_CVPR} and MAML~\cite{finn2017model} with the same backbone and trainging/testing protocol. For our CFA model, we report the results for $N$=1 which only considers spatial compositionality and $N$=64 that imposes both semantic and spatial compositionaliy. As shown in Table~\ref{table1}, our CFA performs better than other methods on the miniImagenet dataset for different sizes of support set.  The result on CUB dataset is shown in Table~\ref{table2}.  Our CFA ($N$=64) outperforms other methods for both 1-shot and 3-shot tasks, and achieves a similar accuracy to ProtoNet for 5-shot task on this dataset. The similar result of our CFA ($N$=64) comparing to ProtoNet  for 5-shot task may be due to that the CUB dataset as a fine-grained bird dataset showing small intra-class variance, thus a larger support set will help to estimate better class center and greatly benefit methods like ProtoNet that rely on distribution estimation.  Moreover, due to the relatively smaller inter-class variance of CUB dataset (all images are birds) comparing to miniImagenet, the benefits of imposing spatial compositionality is limited. Thus the CFA ($N$=1), achieves better accuracy on miniImagenet but performs worse on CUB than these state-of-the-art methods. However, by further incorporating semantic compositionality, our CFA with $N$=64 shows better accuracy on both datasets especially for low-shot cases. The improvement achieved on both dataset shows our method is effective to learn from a few examples for both generic and fine-grained image classification.

\vspace{-5pt}
\subsection{Action Recognition}   
\vspace{-5pt}
\noindent\textbf{Dataset.} We evaluate the performance for few-shot action recognition on two datasets: Kinetics-CMN~\cite{Zhu_2018_ECCV} and Jester~\cite{Jester}. The Kinetics-CMN dataset contains 100 classes with 100 examples for each selected from the Kinetics dataset~\cite{kay2017kinetics}. We evaluate with the splits provided by~\cite{Zhu_2018_ECCV} with 64, 12, 24 non-overlapping classes for base, validation, and novel classes, respectively. The Jester dataset is a hand gesture dataset containing 27 categories of hand gestures with 148,092 video samples in total. In our experiments, we randomly select 1,000  video samples for each hand gesture and then randomly split the  27 classes into 13, 5, 9 non-overlapping classes to be the base, validation, and novel categories, respectively. 

\noindent\textbf{Implementation.} To extract feature maps from the video sequences, we adopt the RGB stream of the two-stream model~\cite{simonyan2014two} with ResNet18~\cite{he2016deep} as backbone. Following the practice in~\cite{Zhu_2018_ECCV},  10 frames are randomly sampled from each video  to be the input sequence for deep CNN. We initialize the backbone with parameters pretrained on the Imagenet dataset~\cite{deng2009imagenet}, then perform episode training for 10 epochs for each dataset. It should be noted that for the action recognition task, our CFA is extended to temporal dimension, and aggregates the feature maps spatio-temporally. The model for the best accuracy is selected with the validation set. As in the image-level few-shot learning task, we also implement recent state-of-the-art methods~\cite{kay2017kinetics,snell2017prototypical,Sung_2018_CVPR} for comparisons. For these methods,
spatio-temporal feature maps are mean pooled to be the video-level representation as in~\cite{simonyan2014two}. All the methods are trained using Adam optimizer with initial learning rate 0.0001 and batchsize 1. During episode testing stage, we randomly sample 20,000 episodes  as in~\cite{Zhu_2018_ECCV} and take the mean accuracy as well as the 95$\%$ confidence intervals as the final results.     
  \begin{table}
  \small
  \centering
    \begin{tabular}{ccccccc}
     \whline
        \vspace{0.015in}
        &CFA(\footnotesize{N=64}) &CFA(\footnotesize{N=1}) &\footnotesize{MatchingNet}\footnotesize{\cite{vinyals2016matching}}  &ProtoNet\footnotesize{\cite{snell2017prototypical}} &\footnotesize{RelationNet}\footnotesize{\cite{Sung_2018_CVPR}} &CMN$^{\dagger}$\footnotesize{\cite{Zhu_2018_ECCV}}\\ 
     \shline
      1-shot &\textbf{69.9 $\pm$ 0.9} &67.7 $\pm$ 0.9 &66.2 $\pm$ 0.9 &55.4 $\pm$ 0.9 &59.1 $\pm$ 0.8 &60.5\\ 
      3-shot &\textbf{80.5 $\pm$ 0.8} &78.1 $\pm$ 0.9 &77.1 $\pm$ 0.8 &76.1 $\pm$ 0.8 &70.7 $\pm$ 0.8 &75.6\\ 
      5-shot &\textbf{83.1 $\pm$ 0.8} &80.9 $\pm$ 0.8 &78.7 $\pm$ 0.7 &81.6 $\pm$ 0.8 &74.4 $\pm$ 0.7 &78.9\\ 
     \whline
     \end{tabular}
    \vspace{-0.0cm}
    \caption{The 5-way few-shot action recognition accuracy (in $\%$) on the Kinetics dataset. The ``$N$'' is the number of semantic subspaces. The results for CMN are  copied from by the original paper~\cite{Zhu_2018_ECCV}.}
   \label{table3}
    \vspace{-0.5cm}
  \end{table}
  \normalsize
  \begin{table}
  \small
  \centering
    \begin{tabular}{cccccc}
     \whline
        \vspace{0.015in}
        &CFA(\footnotesize{N=64}) &CFA(\footnotesize{N=1})  &\small{MatchingNet}\small{\cite{vinyals2016matching}}  &ProtoNet\small{\cite{snell2017prototypical}} &\small{RelationNet}\small{\cite{Sung_2018_CVPR}} \\ 
     \shline
      1-shot &\textbf{69.2 $\pm$ 0.8} &63.6 $\pm$ 0.7 &54.0 $\pm$ 0.7 &51.3 $\pm$ 0.8 &55.3 $\pm$ 0.7 \\ 
      3-shot &\textbf{78.6 $\pm$ 0.6} &69.9 $\pm$ 0.7 &70.2 $\pm$ 0.7 &58.0 $\pm$ 0.8 &64.9 $\pm$ 0.7 \\ 
      5-shot &\textbf{82.3 $\pm$ 0.6} &73.1 $\pm$ 0.7 &73.3 $\pm$ 0.6 &61.7 $\pm$ 0.7 &69.3 $\pm$ 0.6 \\ 
     \whline
     \end{tabular}
    \vspace{-0.0cm}
    \caption{The 5-way few-shot action recognition accuracy (in $\%$) on the Jester dataset.}
   \label{table4}
    \vspace{-0.7cm}
  \end{table}
  \normalsize
  
\noindent\textbf{Results.} We compare our method with several approaches including MatchingNet~\cite{vinyals2016matching}, ProtoNet~\cite{snell2017prototypical}, RelationNet~\cite{Sung_2018_CVPR}, CMN~\cite{Zhu_2018_ECCV}. As presented in Table.~\ref{table3}, our CFA ($N$=64) outperforms other methods for both 1-shot, 3-shot, and 5-shot recognition on the Kinetics-CMN datasets. Given the very large spatio-temporal sample space for videos and the very limited training data (100 videos for each base class) used, it is challenging for deep models to effectively learn a mapping that generalizes well. As a result, the distribution based model ProtoNet performs worst among these methods for small support set like 1-shot task. In contrast, our CFA achieves 69.9$\%$ for 1-shot case, which is nearly 14$\%$ higher than ProtoNet. The large gap shows the effectiveness of spatio-temporal and semantic compositionality imposed by our method. The Jester dataset is a harder dataset for few-shot learning since the inter-class variety for hand gestures is much smaller than generic actions like those in the Kinetics-CMN. As shown in Table~\ref{table4}, previous methods like MatchingNet, ProtoNet and RelationNet drastically drop their accuracy comparing to our CFA ($N$=64). By only consider the spatial bilinear aggregation, rather than applying mean pooling to the spatio-temproal feature maps, our CFA ($N$=1) still achieves better performance than previous methods especially for 1-shot task. By further imposing the semantic compositionality into deep feature, the CFA ($N$=64) achieves a much higher accuracy. This shows that both the semantic and spatial compostionaliy is important for few-shot action recognition, and our CFA is effective to learn to produce discriminative compositional feature in a weakly-supervised way.

\subsection{Method Analysis}

\begin{figure}
\centering
\includegraphics[height=3.2cm] {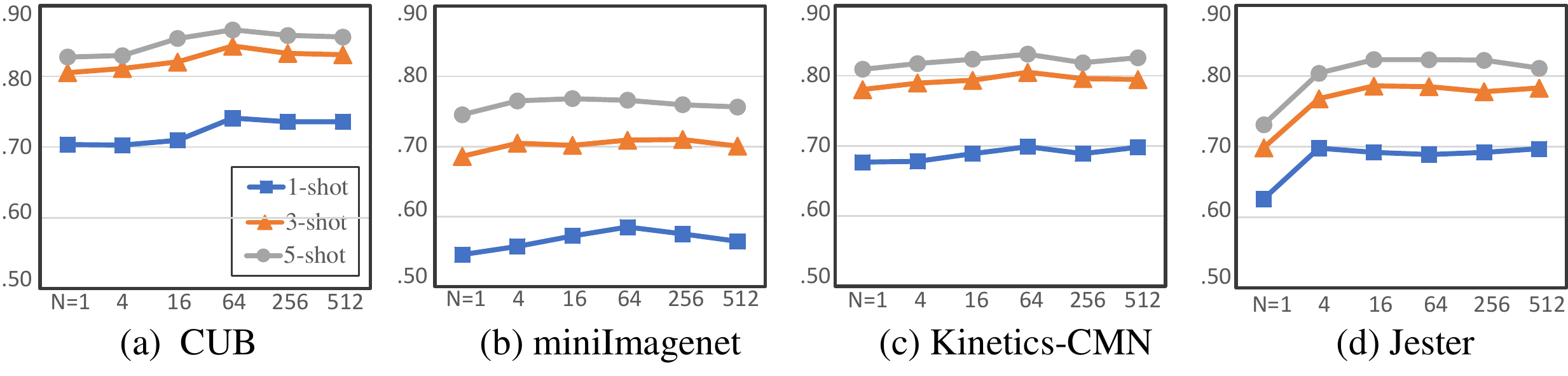}
\vspace{-0.35cm}
\caption{Effect of different numbers of groups ``$N$'' on the four datasets ($\gamma$ = 0.0002).}
\label{fig3}
\vspace{-0.1cm}
\end{figure}
\begin{figure}
\centering
\includegraphics[height=3.2cm] {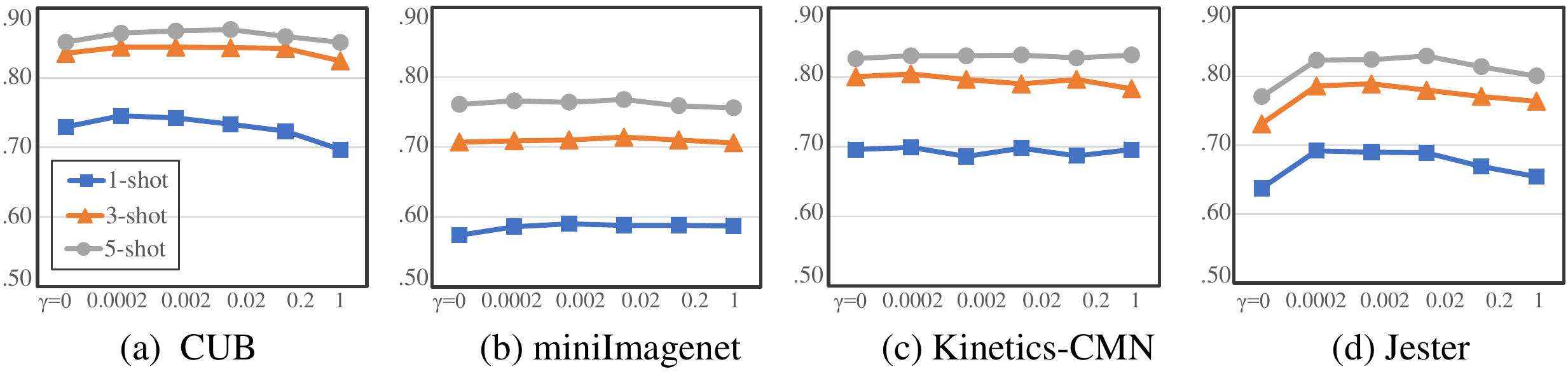}
\vspace{-0.35cm}
\caption{\small Effect of different weights of orthogonality ``$\gamma$'' on the four datasets ($N$ = 64).}
\label{fig4}
\vspace{-0.5cm}
\end{figure}

We first analyse the effect of $N$, which is the number of predefined semantic subspaces. The results for different values of $N$ on the four datasets are presented in Fig.~\ref{fig3}. When incrasing $N$ from 1, the improvements on the miniImagenet and Kinetics-CMN datasets are less obvious than those on the CUB and Jester datasets, indicating that for generic image/action classification the spatial/spatiotemporal compositionality is more effective while for fine-grained classification the semantic compositionality plays a more important role.  Moreover, on both CUB and Jester datasets, the accuracy improves greatly for some values of $N$ (N=64 in Fig. ~\ref{fig3}(a) and N=4 in Fig. ~\ref{fig3}(d)) and then becomes stable. This shows that for fine-grained datasets, there exists a group of optimal semantic attributes that generalize well, and our CFA model can effectively learn to find them with a relatively larger $N$.

\begin{wraptable}{r}{7.5cm}
  \small
  \centering
    \begin{tabular}{p{1.5cm}p{0.95cm}p{0.95cm}p{0.95cm}p{0.95cm}}
     \whline
        \vspace{0.015in}
                                        &\multicolumn{2}{c}{\footnotesize{IMN$\rightarrow$CUB}} &\multicolumn{2}{c}{\footnotesize{KIN$\rightarrow$JSR}}\\ 
                                        &\footnotesize{1-shot}&\footnotesize{5-shot}&\footnotesize{1-shot}&\footnotesize{5-shot}\\ 
     \shline                                  
     \footnotesize{CFA\scriptsize{(N=64)}}        &\footnotesize{\textbf{44.2$\pm$0.8}}&\footnotesize\textbf{{66.0$\pm$0.7}}&\footnotesize{23.9$\pm$0.5}&\footnotesize{28.5$\pm$0.5}\\  
     \footnotesize{CFA\scriptsize{(N=~1)}}         &\footnotesize{39.5$\pm$0.7}&\footnotesize{64.9$\pm$0.7}&\footnotesize{\textbf{24.7$\pm$0.5}}&\footnotesize{\textbf{32.7$\pm$0.6}}\\ 
     \footnotesize{MathingNet}         &\footnotesize{37.4$\pm$0.7}&\footnotesize{55.8$\pm$0.7}&\footnotesize{22.1$\pm$0.5}&\footnotesize{26.9$\pm$0.5}\\ 
     \footnotesize{ProtoNet}           &\footnotesize{37.2$\pm$0.6}&\footnotesize{62.3$\pm$0.7}&\footnotesize{22.8$\pm$0.4}&\footnotesize{27.3$\pm$0.6}\\ 
     \footnotesize{RelationNet}        &\footnotesize{35.9$\pm$0.7}&\footnotesize{56.3$\pm$0.6}&\footnotesize{20.8$\pm$0.3}&\footnotesize{21.9$\pm$0.3}\\ 
     \whline
     \end{tabular}
    \vspace{-0.15cm}
    \caption{\small Cross domain few-shot recognition accuracy (in $\%$). ``\textit{IMN}$\rightarrow$\textit{CUB}'' represents training on miniImagenet and testing on CUB. ``\textit{KIN}$\rightarrow$\textit{JSR}'' means training on Kinetics-CMN and testing with Jester. }
   \label{table5}
    \vspace{-0.3cm}
\end{wraptable}
\normalsize

We also show in Fig.~\ref{fig4} the effect of weights for the orthogonality constraint $\gamma$ in the loss function Equation~\ref{eq5}.    As we can see, on the minImagenet dataset and Kinetics-CMN datatet that contain more generic scenes and objects, the  performance is less sensitive to the value of $\gamma$, since the high inter-class variance and intra-class variance help our CFA to find and summarize meaningful centers. For CUB and Jester datasets, without the orthogonality constraint ($\gamma=0$) it may learn trivial solutions leading to low accuracy. However, a too high $\gamma$ will also harm the performance, as these two datasets have small inter-class and intra-class variance, so forcing the semantic prototypes to be orthogonal to each other may lead to a poorly generalizing representation.   

At last, to evaluate our method's ability for cross domain few-shot learning, we train the model on datasets for generic classification and then test the performance on datasets for fine-grained category recognition. As shown in Table~\ref{table5}, our CFA shows better transfer ability than other methods for both image classification and action recognition. In the first two rows, we compare our CFA with semantic compositionality (N=64) and without compositionality (N=1). For both tasks, by only considering spatial compositionality, CFA (N=1) achieves better performance than previous methods. Further considering attribute compositionality, CFA (N=64) leads to further improvement on image classification tasks, while worse accuracy on video recognition tasks. The decrease for video transfer-learning tasks may be because the Kinetics-CMN dataset, which has only 10,000 videos in total, is too small to learn effective semantic sub-features that generalize well the indoor hand gesture dataset.

\vspace{-5pt}
\section{Conclusion}
\vspace{-5pt}
In this paper, we propose the Compositional Feature Aggregation module as a plugable end-to-end layer for few-shot learning task. By decomposing the feature space into attribute subspaces and applying bilinear local aggregation in each subspace, CFA successfully imposes both spatial and semantic compositionality as a regularization to improve FSL, and produces more discriminative representations. We evaluate our model for both generic and fine-grained image classification and video classification task, and the improvements on all four datasets validate the proposed method's effectiveness.

{\small
\bibliographystyle{abbrv}
\bibliography{ref}
}

\end{document}